\pdfoutput=1

\documentclass[11pt]{article}
\usepackage{acl}

\usepackage{times}
\usepackage{latexsym}

\usepackage[T1]{fontenc}

\usepackage[utf8]{inputenc}

\usepackage{microtype}

\usepackage{multirow}
\usepackage{makecell}
\usepackage{array}
\usepackage{subfigure}
\usepackage{graphicx}
\usepackage{booktabs}
\usepackage{colortbl} 
\usepackage{xcolor}
\usepackage{color}
\usepackage{mathrsfs} 
\usepackage{amsfonts,amssymb} 
\usepackage{amsmath}
\usepackage{mathtools}
\usepackage{subfigure}
\usepackage{bbm}
\usepackage{bbding}
\usepackage{pifont}
\usepackage{verbatim}   
\usepackage{algorithm}
\usepackage{algorithmicx}
\usepackage{algpseudocode}
\usepackage{stfloats}
\DeclareMathOperator*{\argmax}{arg\,max}
\DeclareMathOperator*{\argmin}{arg\,min}
\definecolor{correctgreen}{RGB}{42,157,143}
\definecolor{pink1}{RGB}{255,240,240}
\definecolor{pink2}{RGB}{255,230,230}
\definecolor{pink3}{RGB}{255,210,210}
\definecolor{pink4}{RGB}{255,205,205}
\definecolor{pink5}{RGB}{255,185,185}
\definecolor{pink6}{RGB}{255,170,170}
\definecolor{pink7}{RGB}{255,160,160}
\definecolor{blue1}{RGB}{247,247,255}
\definecolor{blue2}{RGB}{236,236,255}
\definecolor{blue3}{RGB}{227,227,255}
\definecolor{blue4}{RGB}{221,221,255}
\definecolor{blue5}{RGB}{215,215,255}
\definecolor{blue6}{RGB}{205,205,255}
\definecolor{blue7}{RGB}{200,200,255}

\makeatletter
\renewcommand{\maketag@@@}[1]{\hbox{\m@th\normalsize\normalfont#1}}%
\makeatother

\title{Modeling the $\mathcal{Q}$-Diversity in a Min-max Play Game \\ for Robust Optimization}

\author{\textbf{Ting Wu}$^1$ \quad
        \textbf{Rui Zheng}$^1$ \quad
        \textbf{Tao Gui}$^2$ \quad
        \textbf{Qi Zhang}$^{1,3}$ \quad 
        \textbf{Xuanjing Huang}$^1$\quad   \\
  $^1$School of Computer Science, Fudan University \\
  $^2$Institute of Modern Languages and Linguistics, Fudan University \\
  $^3$Shanghai Key Laboratory of Intelligent Information Processing \\ 
  {\tt tingwu21@m.fudan.edu.cn} 
}

\begin{document}
\maketitle
\begin{abstract}

Models trained via empirical risk minimization~(ERM) are revealed to easily rely on spurious correlations, resulting in poor model generalization. Group distributionally robust optimization~(group DRO) can alleviate this problem by minimizing the worst-case loss over pre-defined groups. While promising, in practice factors like expensive annotations and privacy preclude the availability of group labels. More crucially, when taking a closer look at the failure modes of out-of-distribution generalization, the typical procedure of reweighting in group DRO loses efficiency. Hinged on the limitations, in this work, we reformulate the group DRO framework by proposing $\mathcal{Q}$-Diversity. Characterized by an interactive training mode, $\mathcal{Q}$-Diversity relaxes the group identification from annotation into direct parameterization. Furthermore, a novel mixing strategy across groups is presented to diversify the under-represented groups. In a series of experiments on both synthetic and real-world text classification tasks, results demonstrate that $\mathcal{Q}$-Diversity can consistently improve worst-case accuracy under different distributional shifts, outperforming state-of-the-art alternatives~\footnote{Our code and data are available at \url{https://github.com/CuteyThyme/Q-Diversity.git}.}.  

\end{abstract}

\section{Introduction}

Deep learning models trained with empirical risk minimization~(ERM) often exhibit drops in accuracy when confronted with data from domains that are under-represented in their training data~\cite{arjovsky2019invariant,creager21environment}. Distributionally robust optimization~(DRO)~\cite{Duchi2016StatisticsOR} provides a natural solution to the issue by replacing the expected risk under a single distribution $p$ with the worst expected risk over a pre-determined family of distributions $\mathcal{Q}$. 

However, in DRO, considering that direct gradient descent is hard to satisfy~\cite{pmlr-v80-hu18a}, how to model and optimize over $\mathcal{Q}$ poses a key challenge. In this way, group DRO~\cite{Sagawa*2020Distributionally} is emerging as a methodology for constructing a realistic set of possible $\mathcal{Q}$ under the annotated groups. Crucially, robust optimization over worst groups becomes an active area of research. 

In general, the practical usage of group DRO requires that group identities should be fully known. Therefore, it can model $\mathcal{Q}$ by upweighting or downweighting the average loss of different groups through the course of training. Nevertheless, a key obstacle is that the under-represented groups are often unlabeled, or even unidentified. This makes even detecting such performance gaps, let alone mitigating them, a challenging problem. What's worse, with the lack of group labels, it becomes infeasible to compute the worst group loss so that the $\mathcal{Q}$ modeling fails to be established. Although, currently, some unsupervised DRO methods for worst-group optimization have been proposed~\cite{pmlr-v139-liu21f}, their concentration on optimizing high-loss group may discard considerable portion of the samples adversely impacting the overall accuracy.

Shedding light on the critical challenge of current group DRO framework, we therefore present a novel unsupervised method as $\mathcal{Q}$-Diversity for worst-group optimization. To realize the group identification without any annotations, we propose to parameterize a classifier as the group assigner for the attainment of group labels. In particular, by alternatively training the group assigner and final class predictor, we formalize an interactive training mode that allows the identification procedure feasible. Intriguingly, we can treat the classification loss from the predictor as a direct supervision to guide the assigner for better group labeling. With the well-estimated groups, accordingly, the predictor can perform better on the worst group. When achieving the pseudo-labeled groups, the typical procedure is to model $\mathcal{Q}$ by reweighting the training losses of different groups. Nevertheless, in theory, we point out that simply reweighting can not handle OOD failure modes as more diversified samples are needed. Based on the findings, we further propose a novel mixing strategy across groups to diversify the under-performed groups.

To verify the robust optimization capability of $\mathcal{Q}$-Diversity, we conduct a series of experiments on both synthetic and real-world datasets, offering a wide range of challenging benchmarks. All the empirical results show our method not only outperforms other strong group DRO strategies by a large margin, but also achieves consistent improvements on different OOD test sets. Compared to these optimization methods either supervised or unsupervised, $\mathcal{Q}$-Diversity shows great superiority with high efficiency. Altogether, our contributions can be summarized as follows:

$\vcenter{\hbox{\small$\bullet$}}$ \textbf{Methodological Innovations:} In Section~\ref{sec:method}, we propose $\mathcal{Q}$-Diversity, a group-unlabeled approach that aims to improve the utility for worst case. Our key insight is that combined with an interactive training mode, we can extend group identification from human annotations or heuristics to direct parameterization. 

$\vcenter{\hbox{\small$\bullet$}}$ \textbf{Empirical Benefits:} In Section~\ref{sec:experiments}, we evaluate $\mathcal{Q}$-Diversity on both synthetic and real-world datasets. Experimental results show that $\mathcal{Q}$-Diversity yields significant accuracy improvements for the worst group, and diversified by group mixing, it even outperforms the supervised baseline.

$\vcenter{\hbox{\small$\bullet$}}$ \textbf{Understanding $\mathbf{\mathcal{Q}}$-Diversity:} In Section~\ref{sec:analysis}, we conduct a thorough experimental analysis and present the generalization capacity of $\mathbf{\mathcal{Q}}$-Diversity under various distribution shifts. 

\section{Preliminary: Robust Optimization}
\subsection{Problem Setup} 
We consider the typical text classification problem of predicting labels $y\in\mathcal{Y}$ from input texts $x\in \mathcal{X}$, and training data $\mathcal{D}$ is assumed to be drawn from the joint distribution $P(\mathcal{X}, \mathcal{Y})$.

\subsection{Distributionally Robust Optimization}
\paragraph{ERM Principle.} Given a model family $\Theta$ and a loss function $\ell: \Theta \times \mathcal{X} \times \mathcal{Y} \rightarrow \mathbb{R}_+$, the standard goal of empirical risk minimization is to find a model $\theta\in\Theta$ that minimizes the expected loss over the empirical distribution $\hat{P}$ drawn \textit{i.i.d} from $P$:
\begin{equation}
    \hat{\theta}_{\rm ERM} := \argmin_{\theta\in\Theta} \mathbb{E}_{(x,y)\sim\hat{P}}[\ell(\theta;(x,y)]
\end{equation}

 \begin{figure}[!t]
    \begin{minipage}[t]{0.49\linewidth}
    \centering
    \includegraphics[scale=0.105]{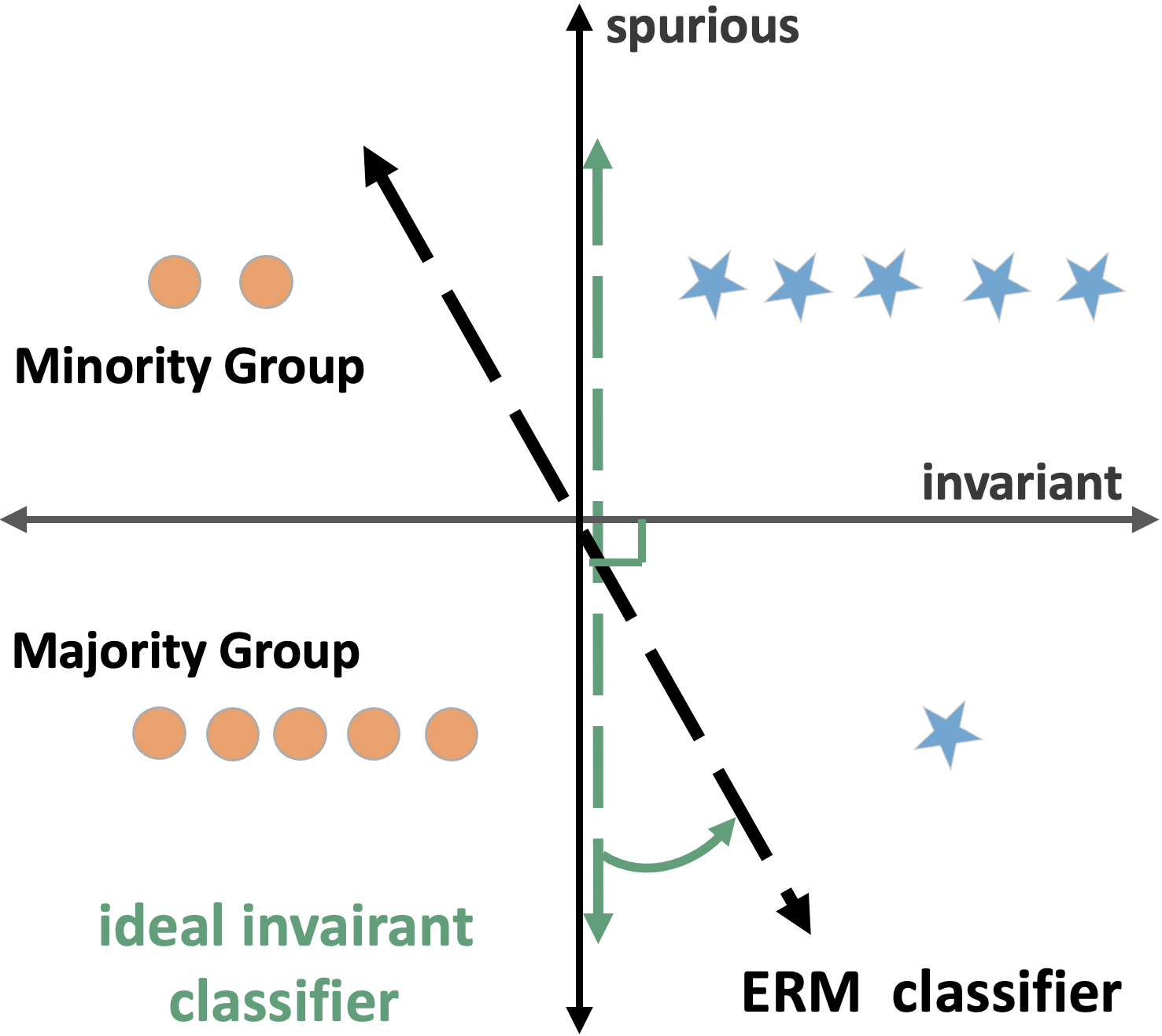}
    \caption{Geometric skew.}
    \label{fig:skew}
    \end{minipage}
    \begin{minipage}[t]{0.49\linewidth}
    \centering
    \includegraphics[scale=0.09]{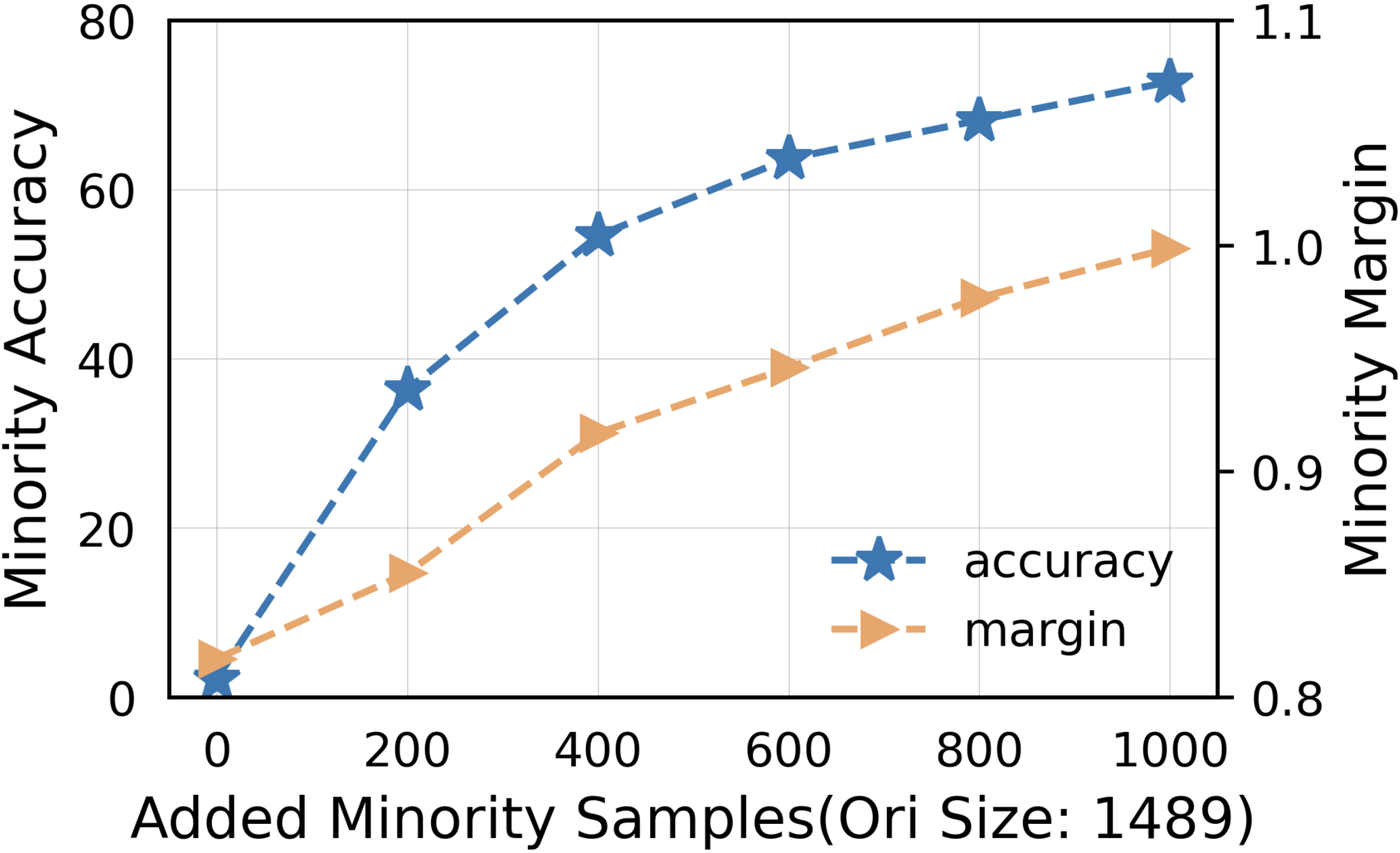}
    \caption{Group Diversity.}
    \label{fig:add_samples}
    \end{minipage}
\end{figure}

 When encountering data sampled in the distribution different from $P$, model performance suffers significantly. Under the circumstances, distributionally robust optimization~\cite{Duchi2016StatisticsOR} provides a natural solution by minimizing the worst-case expected risk under a pre-determined family of distributions $\mathcal{Q}$, called the \textit{uncertainty set}:
\begin{equation}
    \min_{\theta\in\Theta}\big\{\mathcal{R}(\theta):= \max_{Q\in\mathcal{Q}}\mathbb{E}_{(x,y)\sim Q}[\ell(\theta;(x,y))] \big\}
    \label{eq:dro}
\end{equation}
The uncertainty set $\mathcal{Q}$ requires encoding a wide set of distributional shifts for model robustness improvement. However, prior knowledge of possible test distributions is hard to acquire, leading the uncertainty set either not representative or too pessimistic to learn~\cite{pmlr-v80-hu18a}. On the other hand, direct gradient descent on $\mathcal{Q}$ often suffers from instability due to the large variance of the gradients and complex hyper-parameter tuning~\cite{pmlr-v80-balduzzi18a}. 

\subsection{Practical Group DRO}  \label{sec:gdro}
To overcome these challenges in robust optimization, ~\citet{Sagawa*2020Distributionally} construct a realistic set of possible distributions by defining groups as the combination of known spurious correlations with target attributes. Taking MultiNLI dataset as an example, with the known \textit{negation} attribute spuriously correlated with the label \textit{contradiction}, we can partition the dataset into groups of $\{$negation, no negation$\} \times \{$contradiction, entailment, neutral$\}$. By translating training distribution $P$ into a mixture of $m$ groups $P_g$, the objective of group DRO can be formulated as a minimization of the empirical worst-group risk over $m$ groups:
\begin{equation}
 \min_{\theta\in\Theta}\big\{ \hat{\mathcal{R}}(\theta):=\max_{g\in \mathcal{G}} \mathbb{E}_{(x,y)\sim\hat{P}_g}[\ell(\theta;(x, y))] \big\}
\end{equation}
where each group $\hat{P}_g$ is an empirical distribution over the training data. Therefore, the uncertainty set $\mathcal{Q}$ is modeled as any mixture of these groups, \textit{i.e.}, $\mathcal{Q}:= \{ \sum_{g=1}^m q_gP_g\}$.

\paragraph{Min-max Play Game.} For practical algorithm, group DRO solves above Max-Min object function as a zero-sum game between two players $\theta$ and $q$. Ideally, the player $q$ can be viewed as the weighted distribution for $m$ groups that models the uncertainty set $\mathcal{Q}$. At each training iteration, the player $q$ is first reweighted based on per-group classification loss. Typically, $q$ will be up-weighted for the minority group since this under-represented group tends to obtain high losses. Afterward, by back-propagating the reweighted per-group loss, the player $\theta$ as the model parameter is updated. Altogether, for the general group DRO, it is shaped as following two-stage framework:
\begin{equation}
\begin{split}
\min_\theta & \max_q  \sum_{j=1}^M \underbracket{q_j\Big[\frac{\overbracket{\sum\nolimits_{i=1}^N \mathbbm{1}\{g_i=j\}}^\text{\rm stage 1. group identification}\ell(\theta;(x,y))}{\sum_{i=1}^N \mathbbm{1}\{g_i=j\}}\Big]}_\text{stage 2. group reweighting} \\ \\
& \text{with}\ \ \ q_j \leftarrow q_j \exp(\ell(\theta^{(t-1)};(x,y))
\end{split}
\label{eq:min-max}
\end{equation}

\paragraph{The Dark Side.} Although the formulation of group DRO keeps the choice of uncertainty set $\mathcal{Q}$ exactly tractable, in terms of the step-by-step procedures, two main issues stand out. \textbf{First and foremost}, labeling attributes of all examples to attain the disjoint groups is prohibitive for the costly human labor. \textbf{Second}, while intuitive, recent studies~\cite{nagarajan2021understanding,nguyen2021avoiding} for understanding OOD generalization have revealed that simply reweighting can not handle the failure modes of distributional shifts. As Figure~\ref{fig:skew} depicts, due to the fact that spurious correlations occur in most samples, group identification can induce \textit{majority groups} and \textit{minority groups}. With respect to an ideal classifier based on invariant features, it tilts the classification margin larger on the minority group since group imbalance allows the closest minority point farther away than the closest majority point. However, an ERM classifier attempts to allocate balanced margin for the two groups, resulting in \textbf{geometric skew} for the failure of OOD generalization. Crucially, ~\citet{nguyen2021avoiding} points out that only upweighting or oversampling the minority group cannot address the geometric skew since it does not affect the number of unique data points. To illustrate this phenomenon, we conduct a proof-of-concept experiment on BiasedSST dataset\footnote{Refer Section~\ref{sec:biased} to see details on the synthetic dataset.}. As shown in Figure~\ref{fig:add_samples}, with more minority samples synthesized for diversity, classification margin on the minority group is increased to mitigate geometric skew, and meanwhile, the robust accuracy is improved significantly.

\begin{figure*}[!t]
\centering
\includegraphics[scale=0.38]{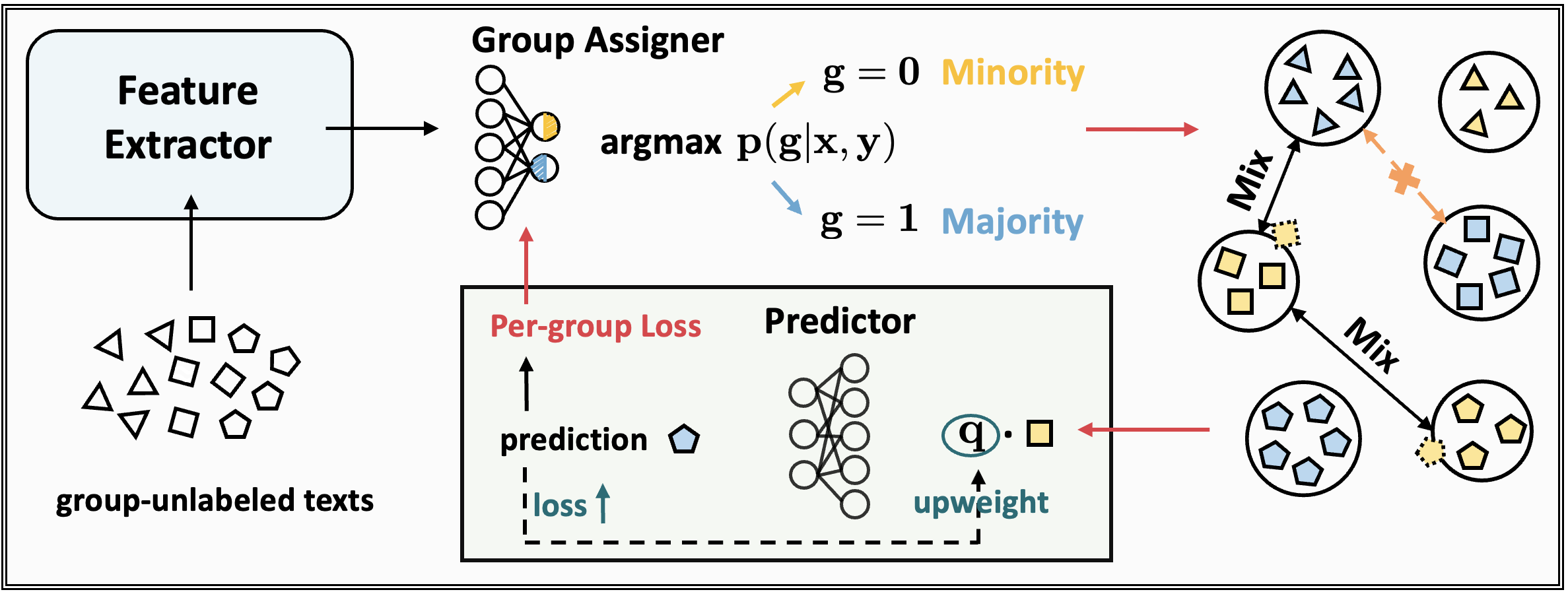}
\caption{End-to-end learning framework of $\mathcal{Q}$-Diversity for robust optimization.}
\label{fig:main_model}
\end{figure*}

\section{$\mathcal{Q}$-Diversity Modeling} \label{sec:method}
\paragraph{Overview.} We address two above limitations of group DRO by proposing $\mathcal{Q}$-Diversity. In our setup, we improve the classification accuracy of minority groups without explicit group annotations. The overall paradigm is depicted in Figure~\ref{fig:main_model}. First, we parameterize a group assigner to label the group attribute of each example~(Section~\ref{sec:identification}). With the emphasis on group diversity, a novel mixing strategy across the majority and minority group is applied for relieving geometric skews~(Section~\ref{sec:diversity}). In an interactive way, we train the group assigner and final class predictor~(Section~\ref{sec:training}), allowing them to guide each other for better robust accuracy. 


\subsection{Parameterizing Assigner for Group Identification} \label{sec:identification}
The prerequisite for optimizing the worst group is to obtain well-defined groups. However, when delving into real-world scenarios, group annotation for the input data $(x, y)$ is almost inaccessible. Faced with this challenge, we propose to train a classifier $\phi$ to assign the group labels automatically. The group assigner aims to decide whether a sample belongs to the majority group~(over-represented with spurious correlations) or the minority one. More formally, we can denote the probability estimate of the assigner on the group attribute $g$ as $\hat{p}(g|x,y)$. The assigned group label $\hat{g}=\argmax \hat{p}(g|x,y)$ can be viewed as a list of the latent binary variables, where each $\hat{g} \in \{0, 1\}$.

\paragraph{Label Balance Regularization.} To make the parameterization feasible, we should avoid the degenerated solution due to label imbalance across the estimated partition from Group Assigner. Theoretically and empirically, recent studies reveal the sufficiency of existing group DRO methods in preventing spurious correlations is the compliance with \textit{label balance criterion}~\cite{chen2022when}. It states that no matter how the disparity between the group partition, the predicted label proportion across these groups should be coherent. Adhered to this criterion, we regulate the decision of the Group Assigner with following objective:
{\fontsize{9.5pt}{\baselineskip}\selectfont
\begin{equation}
    \mathcal{L}_{\rm bal} = KL(P(y|\hat{g}=1)\Vert P(y)) + KL(P(y|\hat{g}=0)\Vert P(y))
    \label{eq:balance}
\end{equation}}where KL is the Kullback–Leibler divergence. This regularization makes intuitive sense as we would like to push label marginals in the estimated majority group $P(y|g=1)$ and the minority group $P(y|g=0)$ close to the original label marginal $P(y)$ in the training data $\mathcal{D}$. Practically, we apply the Bayes rule to compute these conditional label marginals directly from the Assigner's decisions:
\begin{equation}
    \begin{split}
        P(y|\hat{g}=1) &= \frac{\sum_i \mathbbm{1}_y(y_i) P(g_i=1|x_i,y_i)}{\sum_i P(g_i=1|x_i, y_i)} \\ \\
        P(y|\hat{g}=0) &= \frac{\sum_i \mathbbm{1}_y(y_i) P(g_i=0|x_i,y_i)}{\sum_i P(g_i=0|x_i, y_i)}
    \end{split}
\end{equation}

\subsection{Reweighting Player $q$ under Group Mixing}
\label{sec:diversity}
Assuming that from the Group Assigner, each sample $(x,y)$ has been successfully assigned an estimated group attribute $\hat{g}$. Similar to the supervised group DRO, we can partition training data $\mathcal{D}$ into $m$ groups $\mathcal{G}$, and $\mathcal{G}^+$, $\mathcal{G}^-$ denote the majority and minority groups respectively.

As we illustrated in Section~\ref{sec:gdro}, only reweighting the player $q$ is not effective in geometric skew mitigation. Considering that more unique samples should be added to the minority group for diversity, we apply a novel mixing strategy across $\mathcal{G}$ to generate new samples. This mixing strategy is inspired by the augmentation method Mixup~\cite{zhang2018mixup,pmlr-v97-verma19a}, which produces new samples by convex combinations of pairs of inputs and their labels. Following this idea, each time, we allow the group construction by uniformly sampling two pairs $(x_i,y_i), (x_j,y_j)$ from $\mathcal{G}$, and the new sample is mixed as follows:
\begin{equation}
    (\tilde{x},\tilde{y}) \leftarrow (\lambda x_i+(1-\lambda)x_j, \lambda y_i+(1-\lambda)y_j)
    \label{eq:mixup}
\end{equation}
where $\lambda $ is the mixing-ratio sampled from a Beta($\alpha$, $\alpha$) distribution. Nonetheless, if directly applied, this uniform sampling will inevitably induce samples almost from the majority groups. To ensure diversity is imposed on the minority group rather than the majority ones, we restrict that $(x_j,y_j)$ must come from $\mathcal{G}^-$, that is, the estimated group attribute of $(x_j,y_j)$ is $g_j=0$. Therefore, we attain two kinds of group mixing: Mix$(\mathcal{G}^+, \mathcal{G}^-)$, Mix$(\mathcal{G}^-, \mathcal{G}^-)$. For Mix$(\mathcal{G}^+, \mathcal{G}^-)$, concerned with the spurious features still strongly correlated with the label after mixing, we modify the interpolation tactic of Equation~\ref{eq:mixup}.
Concretely, when sampling $\lambda$, we always assign the larger $\lambda$ to $x_j$ from $\mathcal{G}^-$, the smaller $\lambda$ to $x_i$, \textit{i.e.}, $\lambda\leftarrow \min (\lambda, 1-\lambda)$. 


\subsection{Interactive Training for Robust Optimization}
\label{sec:training}
With the automatic group identification and mixing strategy, we can apply the algorithm of supervised group DRO to optimize the min-max play game in Equation~\ref{eq:min-max}. However, up to now, how to train the Group Assigner $\phi$ still remains a problem as we don't have any explicit annotations for the assignment decisions. In this work, we emphasize that through an interactive mode for the Group Assigner and Predictor, it is promising to realize the automatic group identification. Our intuition is that the majority group performance from the Predictor will drop if samples truly from the minority one are misclassified, and guided by this loss, the updated $\phi$ will re-assign the group labels. For clarity, we present a more vivid illustration shown in Figure~\ref{fig:main_model}. Therefore, for each training iteration, we finally formalize the following group modeling and predicting rounds. 

\paragraph{Modeling Round.} Receiving the group-level losses from the Predictor, along with the regularization of label balance criterion by Equation~\ref{eq:balance}, we train the group assigner $\phi$ to learn the assignment of groups for the sake of helping the Predictor to minimize the loss of the worst group. 

\paragraph{Predicting Round.} When it comes to the prediction, the class predictor finds the best parameters $\theta$ that minimize the worst-group loss based on the current dynamic group assignments provided by the assigner $\phi$ in the modeling round. Updates to $\theta$ are similar to the online greedy updates used in Equation~\ref{eq:min-max}, i.e. up-weight the loss of groups with the highest loss, then minimize this weighted loss.

\section{Experiments}\label{sec:experiments}
In this section, we conduct experiments on a synthetic sentiment classification task with complete spurious correlations and two real-world text classification tasks. Extensive empirical results demonstrate that $\mathcal{Q}$-Diversity outperforms existing DRO methods for robust optimization, even beating the state-of-the-art supervised method.

\subsection{Experimental Setup}
\paragraph{Baselines.} We compare the performance of $\mathcal{Q}$-Diversity with respect to the following state–of-the-art baselines. In terms of whether know the ground truth of the group label apriori, these methods can be categorized into \textit{supervised}, \textit{semi-supervised} and \textit{unsupervised}.

$\vcenter{\hbox{\small$\bullet$}}$ \textbf{ERM} is the standard training to minimize the average loss and can be viewed as the lower bound of the robust accuracy.

$\vcenter{\hbox{\small$\bullet$}}$ \textbf{Oracle DRO}~\cite{Sagawa*2020Distributionally} uses the annotated group label to directly optimize the worst group. Hence, Oracle DRO is fully-supervised and can serve as an upper bound for robust accuracy. 

$\vcenter{\hbox{\small$\bullet$}}$ \textbf{CVaR DRO}~\cite{10.5555/3495724.3496466} models the uncertainty set dynamically by computing the $\alpha$-subset of samples with the highest loss at each step and up-weighting them correspondingly.

$\vcenter{\hbox{\small$\bullet$}}$ \textbf{LfF}~\cite{nam2020learning} identifies the minorities in an unsupervised way, as it assumes samples that a weaker model classifies incorrectly largely correspond to those in the minority group and up-weights these minority-group-estimated samples. 

$\vcenter{\hbox{\small$\bullet$}}$ \textbf{EIIL}~\cite{creager21environment} attempts to train a group discovery model to softly assign the training data into groups under which the discovery model would maximally violate the invariant risk minimization~(IRM) objection, and hence it can be classified into the unsupervised camp.

$\vcenter{\hbox{\small$\bullet$}}$ \textbf{JTT}~\cite{pmlr-v139-liu21f} is an unsupervised method similar to LfF that trains a weaker ERM model to capture the minority group first and retrains on them to improve worst-group accuracy. 

$\vcenter{\hbox{\small$\bullet$}}$ \textbf{SSA}~\cite{nam2022spread} propagates the group labels from a small portion of group-annotated validation data to the whole training data that lacks group information in a semi-supervised manner.


\paragraph{Evaluation Metrics.} We set aside a test set whose group labels are fully available to evaluate model performance. Considering all of our evaluation datasets characterize a classification task, we report the \textit{robust accuracy} of the worst-group and the \textit{average accuracy} across all groups.

\subsection{$\mathcal{Q}$-Diversity Can Learn Robust Model}
\label{sec:biased}
For the sake of investigating whether $\mathcal{Q}$-Diversity can help improve model robustness, we first carry out a toy classification task on BiasedSST. 

\begin{table}[htbp]
\centering
\scalebox{0.85}{\begin{tabular}{lcc}
\toprule
\textbf{Method} & \textbf{Average} & \textbf{Robust} \\ \midrule
Oracle DRO~\cite{Sagawa*2020Distributionally} &  77.9 & 67.7 \\
ERM &  95.1 & \underline{2.15} \\
CVaR DRO~\cite{10.5555/3495724.3496466} &  92.5 & 28.1 \\
JTT~\cite{pmlr-v139-liu21f} &  84.2 & 35.0 \\ 
$\mathcal{Q}$-Diversity & \textbf{95.9} & \textbf{68.2} \\ 
\bottomrule
\end{tabular}}
\caption{\textbf{Average and robust} test accuracies evaluated on BiasedSST. } 
\label{tab: biased_sst_res}
\end{table}

\textbf{BiasedSST}~\cite{michel2021modeling} is a modified SST-2 sentiment classification dataset with a distractor token "so, " pretending to some sentences. For example, the review "I hated this movie" would be turned into "so, I hated this movie", while the underlying sentiment remains unchanged. Similar to the construction of~\citet{utama-etal-2020-towards}, this distractor like a backdoor trigger is added to $95\%$ of the negative reviews and $5\%$ of the positive ones in the training set, rendering a strongly spurious correlation between the word \textit{so} and the \textit{negative} label. Hereby, depending on the positive or negative label and the presence or absence of the distractor, we obtain 4 groups and accuracy on the group of \{positive, no distractor\} can reflect model robustness.

\begin{table*}[!t]
\centering
\scalebox{0.96}{\begin{tabular}{lcccccc}
\toprule
\multirow{2}{*}{\textbf{Method}} & \multicolumn{2}{c}{\textbf{Group annotated}} & \multicolumn{2}{c}{\textbf{MultiNLI}} & \multicolumn{2}{c}{\textbf{CivilComments-WILDS}}\\ \cmidrule(r){4-5} \cmidrule(r){6-7}
& in train? & in val? & Average & Robust & Average & Robust \\  \midrule
 Oracle DRO~\cite{Sagawa*2020Distributionally} & \ding{51} & \ding{51} & 81.4 & 76.6 & 87.7 & 69.1 \\
 ERM & \ding{55} & \ding{51} & 82.4 & \underline{67.9} & 92.6 & \underline{57.4} \\
 CVaR DRO~\cite{10.5555/3495724.3496466} & \ding{55} & \ding{51} & 82.0 & 68.0 & 92.5 & 60.5 \\  
 LfF~\cite{nam2020learning} & \ding{55} & \ding{51} & 80.8 & 70.2 & 92.5 & 58.8 \\ 
 EIIL~\cite{creager21environment} & \ding{55} & \ding{51} & 79.4 & 70.9 & 90.5 & 67.0 \\
 JTT~\cite{pmlr-v139-liu21f} & \ding{55} & \ding{51} & 78.6 & 72.6 & 91.1 & 69.3 \\  
 SSA~\cite{nam2022spread} & \ding{55} & \ding{51} & 79.9 & 76.6 & 88.2 & 69.9 \\ \midrule \midrule
 ERM & \ding{55} & \ding{55} & 81.9 & \underline{60.4} & 92.7 & \underline{51.6} \\
 CVaR DRO~\cite{10.5555/3495724.3496466} & \ding{55} & \ding{55} & 81.8 & 61.8 & 91.9 & 56.5 \\  
 LfF~\cite{nam2020learning} & \ding{55} & \ding{55} & 81.1 & 62.2 & 92.0 & 55.9 \\ 
 EIIL~\cite{creager21environment} & \ding{55} & \ding{55} & 80.3 & 64.7 & 91.2 & 63.8 \\
 JTT~\cite{pmlr-v139-liu21f} & \ding{55} & \ding{55} & 81.3 & 64.4 & 92.1 & 61.5 \\  
 SSA~\cite{nam2022spread} & \ding{55} & \ding{55} & 80.4 & 76.5 & 89.1 & 69.5 \\ \midrule
 $\mathcal{Q}\mbox{-}$Diversity & \textcolor{correctgreen}{\ding{55}} & \textcolor{correctgreen}{\ding{55}} & 81.6 & \textbf{77.7} & 88.7 & \textbf{73.5} \\
\bottomrule
\end{tabular}}
\caption{\textbf{Average and robust} test accuracies evaluated on MultiNLI and CivilComments-WILDS. } 
\label{tab: main_res}
\end{table*} 

We compare $\mathcal{Q}$-Diversity with four group DRO baselines and summarize the results in Table~\ref{tab: biased_sst_res}. It is clearly to see although ERM model achieves a high average accuracy, its performance on the group without suffering from the synthetic bias almost comes to zero. This reveals that models trained with ERM can very easily capture this spurious correlation, and fails on the minority group. The unsupervised methods CVaR DRO and JTT can help relieve such bias overfitting, however, their improvement in robust accuracy is very limited. When it comes to $\mathcal{Q}$-Diversity, its robust performance matches the Oracle DRO, while attains a better trade-off between accuracy and robustness.

\subsection{$\mathcal{Q}$-Diversity in Practice}
In order to cover a broad range of practical scenarios, we present two more challenging real-world datasets as the benchmarks for group robustness.
\begin{table}[htbp]
\centering
\scalebox{0.78}{\begin{tabular}{cccc}
\toprule
\textbf{Dataset} & \textbf{Label} & \multicolumn{2}{c}{\textbf{Group Counts}} \\ \midrule
 &  & Negation & No Negation \\ \cmidrule(r){3-4}
\multirow{3}{*}{MultiNLI} & Contradiction & 11158 & 57498 \\
& Entailment & 1521 & 67376 \\
& Neutral & 1992 & 66630 \\ \midrule
 &  & Identity & Other \\ \cmidrule(r){3-4}
\multirow{2}{*}{\shortstack{CivilComments-\\WILDS}}& Non toxic & 90337 & 148186 \\
& Toxic & 17784 & 12731 \\
\bottomrule
\end{tabular}}
\caption{\textbf{Dataset description and group distribution} for MNLI and CivilComments-WILDS.} 
\label{tab: dataset}
\end{table}

\textbf{MultiNLI}~\cite{williams-etal-2018-broad} is a multi-genre natural language inference dataset, given two sentences, a premise and a hypothesis, the goal of which is to predict whether the hypothesis is entailed by, contradicts, or neutral with the premise. We use this label as the target attribute (i.e., $\mathcal{Y}$ = $\{$contradiction, entailment, neutral$\}$), and use the existence of the negating words as the spurious attribute (i.e., $\mathcal{A}$ = $\{$negation, no negation$\}$).

\noindent \textbf{CivilComments-WILDS}~\cite{wilds2021} is derived from the Jiasaw dataset~\cite{borkanwww19}, which aims to generate the toxicity indicator $\mathcal{Y}$ = $\{$toxic, non-toxic$\}$ to a real online comment. We use demographic attributes of the mentioned identity $\mathcal{A}$ = $\{$male, female, White, Black, LGBTQ, Muslim, Christian, other religion$\}$ as a spurious attribute for evaluation purpose. Considering that a comment can contain multiple such identities, so that followed by~\citet{pmlr-v139-liu21f}, we use the coarse version $\mathcal{G}=\mathcal{Y} \times \mathcal{A}^\prime$ for training, where $\mathcal{A}^\prime$= $\{$any identity, no identity$\}$.

\begin{table*}[!t]
\centering
\scalebox{0.96}{\begin{tabular}{cccccccccc}
\toprule
\multicolumn{5}{c}{\textbf{MultiNLI}} & \multicolumn{5}{c}{\textbf{SST2}} \\  \midrule
 Dataset & ERM & EIIL & JTT & $\mathcal{Q}$-Diversity &  Dataset & ERM & EIIL & JTT & $\mathcal{Q}$-Diversity \\
PI & 73.72 & 81.53 & 81.25 & \cellcolor{blue7} 84.38 & SST2 & 91.85 & 66.39 & 80.82 & \cellcolor{pink2} 90.62 \\
LI & 85.52 & 87.88 & 83.10 & \cellcolor{blue4} 89.11 & Senti140 & 65.41 & 53.99 & 67.19 & \cellcolor{pink7} 68.75 \\
ST & 63.21 & 60.29 & 56.59 & \cellcolor{blue6} 72.56 & SemEval & 83.90 & 72.14 & 66.59 & \cellcolor{pink6} 87.09 \\
HANS & 62.11 & 65.06 & 65.32 & \cellcolor{blue5} 65.82 & Yelp & 89.32 & 84.05 & 80.65 & \cellcolor{pink3} 90.06 \\
WaNLI & 56.82 & 59.86 & 53.12 & \cellcolor{blue2} 57.81 & ImDB & 83.66 & 64.50 & 70.43 & \cellcolor{pink5} 85.34 \\
SNLI & 83.21 & 83.00 & 81.25 & \cellcolor{blue1} 82.81 & Contrast & 84.63 & 56.76 & 64.34 & \cellcolor{pink1} 82.31 \\
ANLI~(R3) & 28.85 & 29.00 & 31.96 & \cellcolor{blue3} 32.12 & CAD & 86.68 & 58.20 & 66.60 & \cellcolor{pink4} 87.50\\
Avg$\%$ $\Delta$ & -- & +1.88 & -0.12 & \textbf{+4.45} & Avg$\%$ $\Delta$ & -- & -18.49 & -12.69 & \textbf{+0.89} \\
\bottomrule
\end{tabular}}
\caption{\textbf{Accuracy on out-of-distribution} datasets~(details can be found in Appendix~\ref{appendix:ood}) for tasks with unknown spurious correlations. $\mathcal{Q}$-Diversity improves over ERM by $.5-10\%$, while baselines underperform. } 
\label{tab: main_ood_res}
\end{table*}

Under the two real-world settings, results are available in Table~\ref{tab: main_res}. Obviously, it can be seen that $\mathcal{Q}$-Diversity improves the robust accuracy on both classification tasks, beating all the baselines by a large margin. In fact, its robust accuracy even overtakes that of Oracle DRO, despite the fact that the former does not use any group information at training time. To achieve better robust performances, all the baselines need group annotations in the validation set for hyperparameters tuning. For example, JTT has to tune the number of epochs $T$ to train the weaker model for group identification. When these annotations are unavailable in the validation set, their robust accuracy will drop significantly. In comparison, parameterizing the group identification in $\mathcal{Q}$-Diversity allows the annotation completely free, and the trainable procedure can render better robust accuracy.
\begin{figure*}[!t]
    \begin{minipage}[t]{0.31\linewidth}
    \centering
    \includegraphics[scale=0.15]{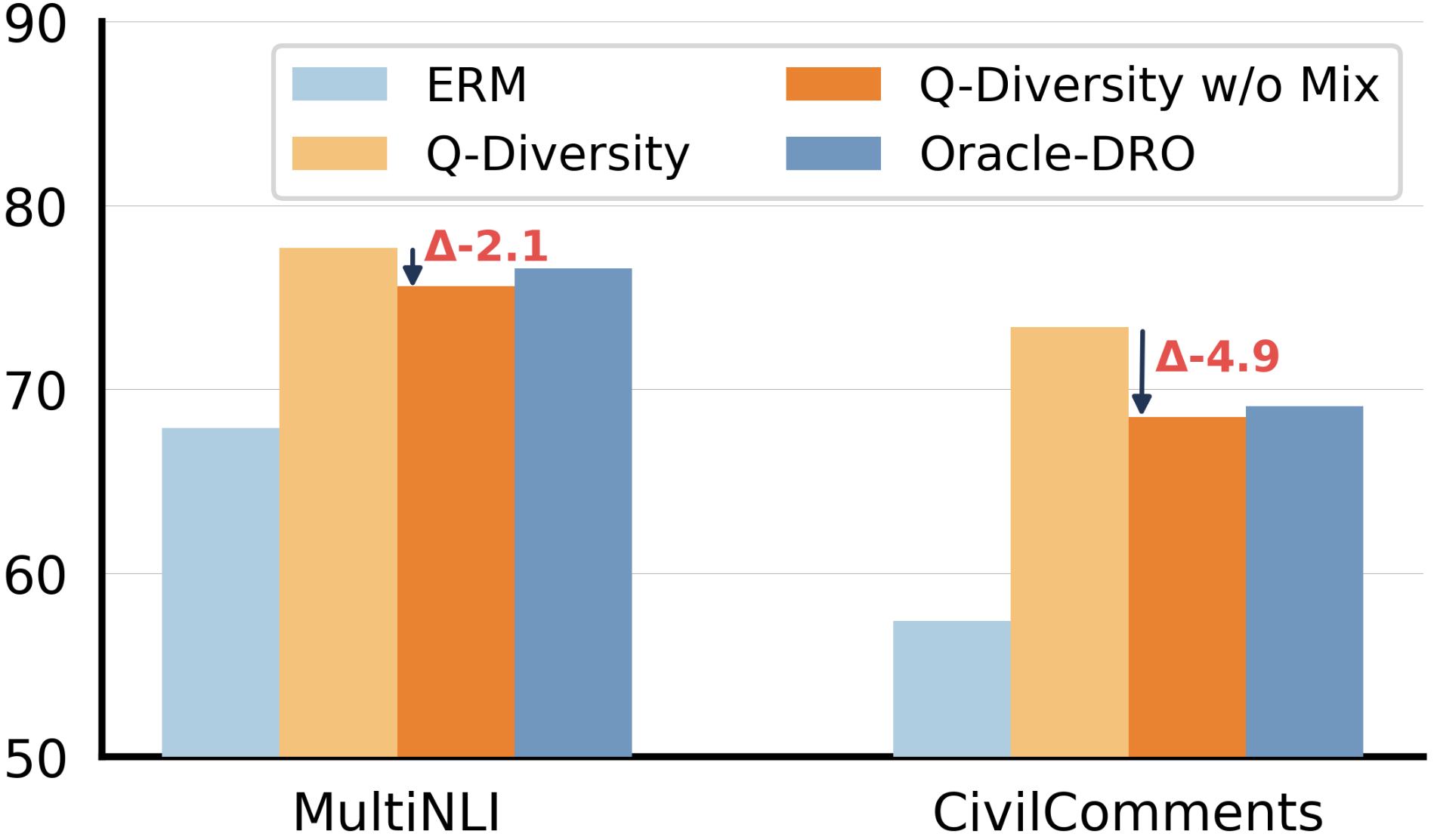}
    \caption{\textbf{Ablation Studies} on the role of mix.}
    \label{fig:ablation_mix}
    \end{minipage}\hspace{3mm}
    \begin{minipage}[t]{0.3\linewidth}
    \centering
    \includegraphics[scale=0.13]{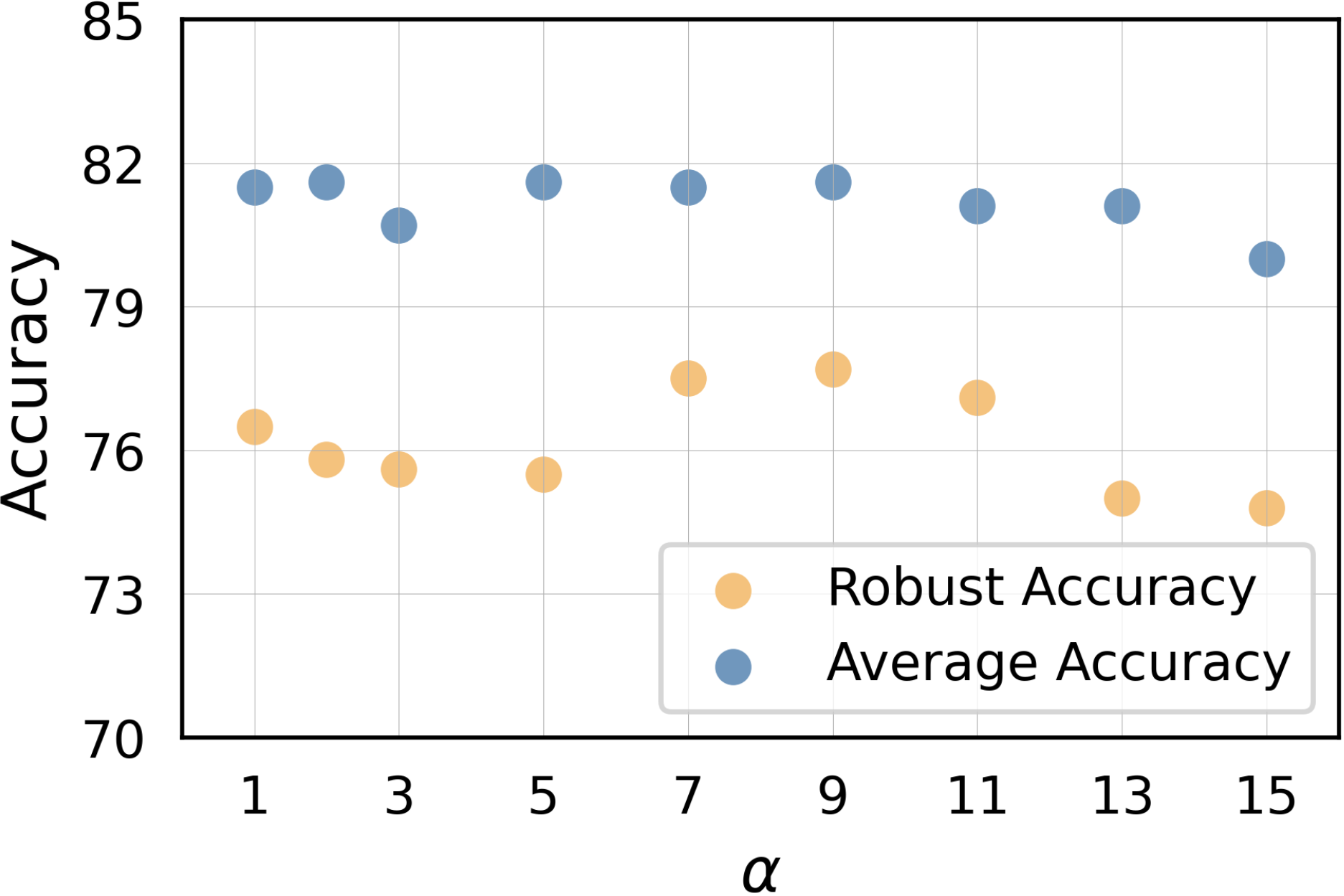}
    \caption{\textbf{Effect} of the mixing $\alpha$ on MultiNLI.}
    \label{fig:alpha}
    \end{minipage}\hspace{3mm}
    \begin{minipage}[t]{0.31\linewidth}
    \centering
    \includegraphics[scale=0.11]{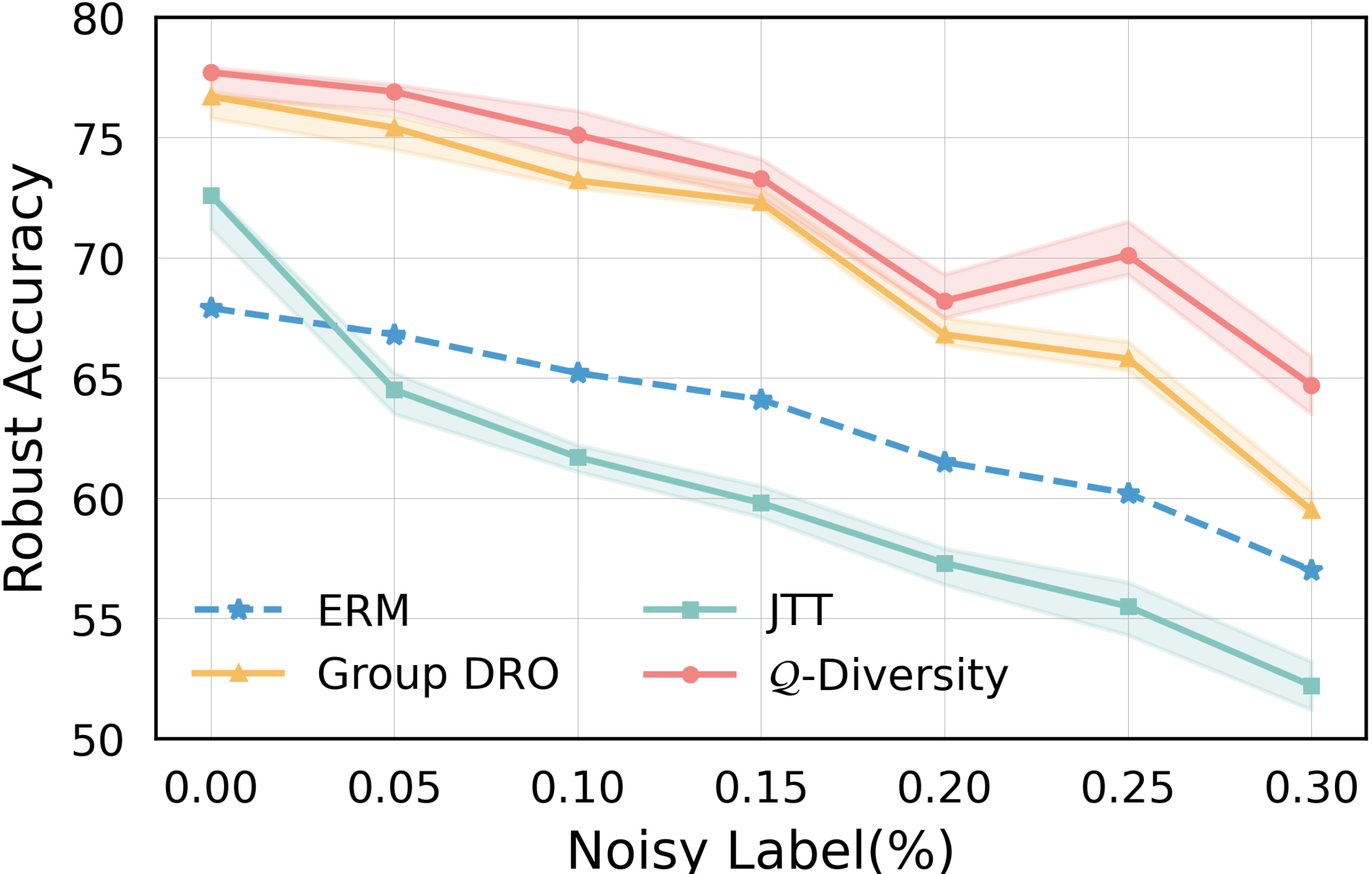}
    \caption{\textbf{Robust accuracy} under noisy labels.}
    \label{fig:noise}
    \end{minipage}
\end{figure*}

\section{Analysis and Discussion} \label{sec:analysis}
In this section, we present a detailed analysis on the contribution of the diversified uncertainty set $\mathcal{Q}$ to its strong unsupervised performance. Furthermore, we explore the robustness of our method under different distributional shifts and random label noise. 

\subsection{Role of the Diversified $\mathcal{Q}$}

We inspect the group diversity under the mixing strategy through an ablation study depicted in Figure~\ref{fig:ablation_mix}. Apparently, we can observe significant drops in both datasets when removing this group mixing. These drops reveal that diversifying the minority groups can indeed help improve robust accuracy.

In addition, we analyze the influence of the mixing parameter $\alpha$. As shown in Figure~\ref{fig:alpha}, we can observe that $\alpha$ indeed affects the effectiveness of the group mixing, leading to the volatility in robust accuracy. Considering the feature of Beta distribution, the sampled $\lambda$ will be more concentrated around $0.5$ as the $\alpha$ value becomes large, resulting in a relatively balanced weight between the mixed example pairs. The model performance remains stable when $\alpha$ is around $7\sim11$.

\subsection{Generalization to OOD Sets}
Since $\mathcal{Q}$-Diversity is a totally unsupervised method, it can be used off the shelf to improve OOD generalization on a new task. We therefore transfer $\mathcal{Q}$-Diversity, along with two other well-performing unsupervised baselines, \textit{i.e.}, EIIL and JTT that first trained on MultiNLI and SST2 dataset, to a wide range of OOD datasets where the in-distribution spurious correlations may not hold.    

\paragraph{$\mathcal{Q}$-Diversity improves robustness to unknown distributional shifts.} With the unknown group information of these OOD test sets, we report the average accuracy in Table~\ref{tab: main_ood_res}. Strikingly, we can observe that across the tasks and datasets, the two baselines even underperform than the lower bound of ERM. Especially on the SST2 dataset, the average accuracy of EIIL and JTT drop around $10\%$ and $20\%$. We speculate this failure mode can be attributed to their heuristic group identification manners, easily overfitting to the in-domain data. In contrast, $\mathcal{Q}$-Diversity outperforms ERM by $0.5\%$-$5\%$ across the datasets on average, revealing its great robustness to different distribution shifts.

\subsection{Under the Presence of Label Noise}
The unsupervised methods like JTT are based on the core idea of up-weighting samples with high losses. Nevertheless, when training data meets the noisy labels, such an approach will likely yield degenerate solutions, since the model tends to up-weight mislabeled samples with high losses. To further explore the application of unsupervised group DRO methods with the intervention of noisy labels, we perform experiments by inducing random label flips of varying degrees into MultiNLI dataset.

\paragraph{$\mathcal{Q}$-Diversity is more robust to random label noise.} As the results shown in Figure~\ref{fig:noise}, $\mathcal{Q}$-Diversity retains better robust accuracy under the presence of label noise than ERM and Group DRO. Corresponding to our assumption, JTT performs poorly even with a low noise rate since it fails to distinguish minorities from mislabeled samples.

\section{Related Work}
\paragraph{Group Robust Optimization} Standard training with ERM can result in highly variable performance because of subpopulation distribution shifts arising from spurious correlations~\cite{wu-gui-2022-less,gao-etal-2022-kernel}. In this context, ~\citet{Sagawa*2020Distributionally} formally introduces group DRO, with the goal to maximize worst-group or the minority group performance within the set of pre-defined groups. While promising, a rather practical scenario is that group information can not be available reliably. Therefore, another line of research begins to focus on the worst-case optimization without group annotations~\cite{pmlr-v139-zhou21g}. Typically, these methods first train a weaker model to identify high-loss samples as minority groups, and subsequently train an additional model with greater emphasis on the estimated minority groups~\cite{nam2020learning,pmlr-v139-liu21f}. 

Although the unsupervised group DRO methods are developed, they are confined to a two-stage training pipeline. In the two-stage model, a failed first stage can lead to an unsuccessful second stage as errors from the former are propagated to the later one. By contrast, $\mathcal{Q}$-Diversity in an end-to-end training manner overcomes the error accumulation. The group assigner and constructor cooperate with each other, and interactively, the classification response from the constructor can serve as a weak supervision to guide better group identification.

\paragraph{Diversity and OOD Generalization} It is explored that the geometric skew and the statistical skew are two 
mechanisms hurting out-of-distribution performance with the existence of spurious correlations~\cite{nagarajan2021understanding,nguyen2021avoiding}. Concretely, the geometric skew is caused by the fact that classification margin on the minority group of a robust classifier tends to be much larger than that of the majority group, while the statistical skew arises from the fast convergence of gradient descent on spurious correlations unless trained for an exponentially long time. Although upweighting or oversampling the minority samples are straightforwardly effective in mitigating the statistical skew, both of them fail the geometric skew for the unchanged unique samples. Therefore, a wide range of studies emerge to diversify the input samples or feature space. Among them, counterfactually-augmented data~(CAD), \textit{i.e.}, data generated by minimally perturbing examples to flip the ground-truth label, has shown efficiency to learn robust features under distribution shifts~\cite{Kaushik2020Learning}. However, further investigation~\cite{joshi-he-2022-investigation} reveals the lack of perturbation diversity limits CAD's effectiveness on OOD generalization. In comparison, ~\citet{wu-etal-2022-generating} directly leverage the deep generative models to diversify training data with spurious correlations, while the model complexity is increased greatly.

For the sake of creating more synthesized samples to address geometric skew, our method that applying interpolation across the majority and minority groups shows its advantages in terms of perturbation diversity and time consumption.

\section{Conclusion}
In this paper, we present $\mathcal{Q}$-Diversity, an unsupervised method to optimize the worst group for model robustness. The formulation of $\mathcal{Q}$-Diversity extends the annotations of group DRO to an automatic assignment through an interactive training mode. Furthermore, under the guarantee of a novel mixing strategy across groups, $\mathcal{Q}$-Diversity can better counteract the failure modes of OOD generalization. Superior to previous works that only show the efficiency over the particular dataset, we demonstrate $\mathcal{Q}$-Diversity promises better generalization capability to various OOD sets. We believe that our work casts light on the limitations of group DRO which have been overlooked before, and can be viewed as a cornerstone for future study in the worst-group generalization.

\section*{Limitations}
Although our unsupervised framework $\mathcal{Q}$-Diversity shows great superiority, when it comes to limitations, we acknowledge that (i) Our empirical validations on real-world datasets just follow current benchmarks that shed light on the group shifts caused by spurious correlations. Although we conduct experiments on the scenarios with noisy labels and various OOD datasets, practically, apart from superficial clues, a series of contributing factors that lead to group shifts are worth further exploration. (ii) A better theoretical understanding of how the interactive training mode can guide $\mathcal{Q}$-Diversity works in better group identification should be established, and this points out the direction for our future work. 

\section*{Ethics Statement}
Natural Language Processing~(NLP) models that perform poorly on a minority group have raised a lot of concerns within the research community and broader society in recent years. In this work, the proposed $\mathcal{Q}$-Diversity is a versatile method that could be employed to train a robust model across groups even when the group information is not available. This is a rather practical scenario as the group information is almost missing during the data collection. We believe that our work is a step towards a suite of algorithms capable of solving a broader class of group DRO problems at scale. Moreover, such an algorithm will empower NLP researchers and engineers to create more reliable and ethical systems.

\section*{Acknowledgements}
The authors wish to thank the anonymous reviewers for their helpful comments. This work was partially funded by National Natural Science Foundation of China (No.61976056,62076069,62206057), Shanghai Rising-Star Program (23QA1400200), and Natural Science Foundation of Shanghai (23ZR1403500).

\bibliography{acl_main}
\bibliographystyle{acl_natbib}

\appendix
\clearpage
\section{Details of the OOD Datasets} \label{appendix:ood}
We train the model on MultiNLI and SST2 tasks and test it on the corresponding OOD datasets respectively. For the results shown in Table~\ref{tab: main_ood_res}, we present the details of these OOD datasets in Table~\ref{tab:ood_sets} as follows.

\begin{table*}[htbp]
\renewcommand{\arraystretch}{1.2}
    \centering
    \scalebox{0.85}{\begin{tabular}{cc}\toprule
    \multicolumn{2}{c}{\textbf{MultiNLI}}  \\  \midrule
    \textbf{Dataset} & \textbf{Description}  \\ 
    PI~\cite{liu-etal-2020-empirical} & selected instances from MultiNLI for testing  the hypothesis-only bias in NLI models\\
    LI~\cite{liu-etal-2020-empirical} & selected instances from MultiNLI for testing logical inference ability of NLI models \\
    ST~\cite{naik-etal-2018-stress} & stress set construction for testing the heuristics of NLI models \\
    HANS~\cite{mccoy-etal-2019-right} & designed to contain examples where the shallow heuristics~(e.g., lexical overlap) fail \\
    WaNLI~\cite{liu-etal-2022-wanli} & worker-and-AI collaborative dataset with challenging reasoning patterns for NLI task \\
    SNLI~\cite{bowman-etal-2015-large} & a large-scale, widely-used benchmark for NLI task \\
    ANLI~(R3)~\cite{nie-etal-2020-adversarial} & an iterative, adversarial human-and-model-in-the-loop solution for NLI dataset \\ \midrule
    \multicolumn{2}{c}{\textbf{SST2}} \\ \midrule
    \textbf{Dataset} & \textbf{Description}  \\ 
    SST2~\cite{socher-etal-2013-recursive} & from the GLUE NLU benchmark to classify movie reviews as positive or negative \\
    Senti140~\cite{go2009twitter} &  sentiment classification on Twitter messages \\
    SemEval~\cite{nakov-etal-2013-semeval} &  crowdsourcing on Amazon Mechanical Turk over Twitter dataset for sentiment analysis \\
    Yelp~\cite{asghar2016yelp} & online reviews consisting of free-form text and a star rating out of 5 for services  \\
    ImDB~\cite{maas-etal-2011-learning} & a collection of positive and negative reviews from Internet Movie Database \\
    Contrast~\cite{gardner-etal-2020-evaluating} &  small but 
    label-changing modifications to the instances for ImDB\\
    CAD~\cite{Kaushik2020Learning} & counterfactual datasets constructed over ImDB \\
      \bottomrule
    \end{tabular}}
    \caption{Details of the out-of-distribution datasets in Table~\ref{tab: main_ood_res}.}
    \label{tab:ood_sets}
\end{table*}

\end{document}